# Investigating the Effect of *k*-NN Preprocessing on Developing Graph Neural Networks: A Fairness-Based Perspective

Nikolaos Zafeiropoulos
*Department of Cultural Technology and Communication*
*University of the Aegean*
Mytilene, Greece
nzaf@aegean.gr

Emmanouil Mavrikos
*Department of Cultural Technology and Communication*
*University of the Aegean*
Mytilene, Greece
emmmavrikos@aegean.gr

George E. Tsekouras
*Department of Cultural Technology and Communication*
*University of the Aegean*
Mytilene, Greece
gtsek@aegean.gr

***Abstract*—In this paper, a methodology to design fair graph convolutional neural networks (GCNs) is developed and tested over several application data sets. The graphs that are used as inputs to the network are constructed by a *k*-nearest neighbor-based preprocessing procedure, while fairness issues are considered in terms of the equalized odds criterion. To effectively incorporate the above heterogenous information, the equalized odds criterion is directly embedded into the model's optimization objective through an additional fairness-driven loss functional term. The proposed methodology investigates how varying the neighborhood size in the *k*-NN algorithm during graph construction influences both the classification performance and the fairness of the resulting models. Extensive experimentation is conducted on three real-world tabular datasets with known biases, evaluating the interplay between graph structure and fairness enforcement. The results demonstrate that the choice of the value of the parameter *k* critically impacts the performance trends, either steadily improving or peaking at intermediate values depending on dataset characteristics, while the application of fairness constraints significantly mitigates disparities in false positive and false negative rates across groups defined by the protected variable at hand, without incurring major sacrifices in overall accuracy. This study highlights the importance of jointly optimizing the graph construction process and fairness objectives in GCN-based learning, providing a systematic approach toward building more equitable and effective graph-based models.**

## I. Introduction

In recent years, Graph Neural Networks (GNNs) have emerged as a powerful class of models capable of learning from data represented in graph structures, where relations between entities are as important as the entities themselves [1]. So far, several approaches based on different graph-based representation techniques have been put forth [1, 2]. In addition, the utilization of deep learning models showed a positive impact due to their inherent ability to process graph-structured data directly (e.g., using graph embeddings) [2].

A well-known type of GNNs is the Graph Convolutional Networks (GCNs). GCNs have attracted special attention due to their ability to learn expressive node representations by collecting information coming from neighboring nodes. As a result, their applications span diverse areas such as recommendation systems, social networks, and biology [3]. Despite their promising capabilities, the performance of GCNs strongly depends on the underlying graph structure, which is often designed and approximated before training. That strategy raises important questions about the quality of learning representations, especially when the graph is not naturally given. In addition, a critical issue is related to the development of fair (i.e., debiased) GNNs.

Fairness in machine learning (ML) has become a central design direction, especially when socially sensitive domains (e.g., criminal justice, hiring, financial sector, institution entry, etc.) are involved [4]. Trained ML models on historical data often inherit and grow prejudices existed in the data, leading to discriminatory decision-making against certain demographic groups [4, 5]. A group that is favored by the ML-model decisions is called *privileged* or *non-sensitive*, whereas in the opposite case, it is called *unprivileged* or *sensitive*. The privileged and unprivileged groups are defined by a binary attribute called *protected attribute*. To anticipate such kind of discrimination and counterbalance its effects, fairness criteria are involved in model's developing process with the ultimate purpose of mitigating the corresponding bias that exists in the training data and providing fairer ML-based decision-making regarding the unprivileged group [6, 7]. Most current measures of fairness, such Equalized Odds (EO) and demographic parity (DP), purportedly consider discrete sensitive characteristics like gender, race, etc. In particular, the EO criterion has been widely involved within the fair ML perspective because it requires simultaneous minimization of type I and type II errors [8]. Minimization of type I implies that the false positive rates (FPRs) between the privileged and unprivileged groups must be similar. Relationally, minimization of type II error assumes that the false negative rates (FNRs) must be similar across the above-mentioned groups.

Regarding GCNs, maintaining fairness can be particularly challenging because the relation dependence between nodes might promote and increase prejudice, which in turn can be propagated through the network [9]. In addition, developing fair GCN models constitutes an open research question, especially when the main target is to perform bias mitigation without compromising the resulting model's accuracy [10]. To address these issues, recent works have proposed integrating fairness-aware constraints into GNN architectures. For example, Ge et al. [11] proposed an impartiality-individual GNN framework that involves EO obstacles while other similar attempts studied how structural bias contributes to unfair decisions [12]. The above models either assume a fixed debiased graph structure or apply post-hoc (i.e., after the model has been created) fairness-based analysis [13]. However, this strategy may not be effective in canceling the correlation effects between data attributes that favor discriminative decision-making, resulting in inefficient model capabilities regarding bias mitigation.

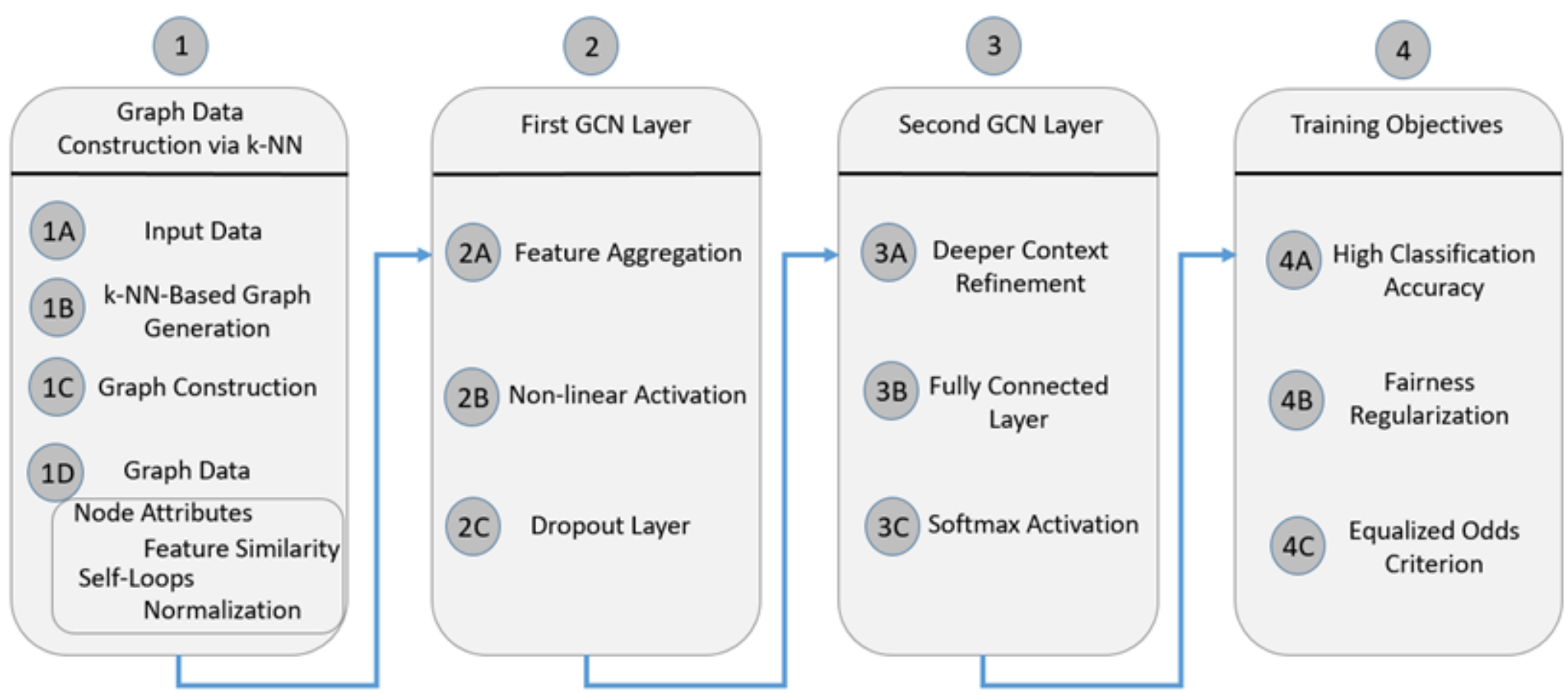

Fig. 1. The overall methodological procedure.

Such correlations could be present in the available dataset; they can be encoded in the graph structure and finally passed through the network. To deal with the above difficulties, this paper focuses on developing fair GCNs using an in-processing technique to mitigate bias, where the basic structure of the network is carried out in terms of *k*-NN procedure. In that sense, the main contributions of the current endeavor are summarized as follows:

(a) The *k*-nearest neighbor (*k*-NN) algorithm is used to build the network's structure, where the number of edges per node is determined by the value of the parameter *k*, enabling a regulated and repeatable graph generation process.

(b) The EO is used as constraint to modify the objective function of the GCN and to guide the learning process in obtaining fair decision-making processes.

(c) The effect of the parameter *k* on the above-mentioned fair decision-making is quantified in terms of statistical inference, with the ultimate purpose of selecting an appropriate value of that parameter given a certain data set.

The remainder of this paper is organized as follows: Section II presents the methodology, including data preprocessing, graph construction, and model training. Section III outlines the simulation study, reporting on performance and fairness outcomes across the datasets. Along with the statistical analysis. Finally, Section IV concludes with key findings and outlines directions for future research.

## II. Methodology

### A. Architecture

The overall pipeline is depicted in Fig. 1. The methodology integrates *k*-NN based graph construction, node-level convolutional learning, and fairness-aware training into a unified pipeline. The process begins with the input data (1A), a tabular dataset comprising both predictive features and the protected attribute. Since the data lacks inherent graph structure, it undergoes *k*-NN preprocessing (1B), which identifies the nearest neighbors for each node based on feature similarity, forming the edges of the graph. This leads to the graph construction phase (1C), which incorporates essential components such as node attributes, similarity-based edges, self-loops, and normalization (1D) to produce a learning-ready graph representation.

Learning begins in the first GCN layer, where the model performs feature aggregation (2A) across neighboring nodes. This is followed by a non-linear activation function (2B) (e.g., ReLU) to enable the model to capture complex structural patterns, and a dropout layer (2C) that serves as a regularization mechanism to prevent overfitting. The second GCN layer carries out deeper context refinement (3A) by further propagating and transforming the node embeddings. These are then passed through a fully connected layer (3B) and finally elaborated by softmax activation (3C) to produce probabilistic class predictions. The training objective of the architecture comprises two parallel goals: achieving high classification accuracy (4A) and maintaining fairness (4B), enforced via the EO criterion (4C) to reduce disparities in error rates across sensitive demographic groups. This layered architectural design provides a principled foundation for transforming unstructured tabular data into a graph-learning task that is not only performance-oriented but also fairness-aware. By sequentially applying structure-inducing transformations and leveraging message passing through GCN layers, the model effectively learns both local and global data patterns. More importantly, the inclusion of EO-based regularization ensures that the model's predictions do not disproportionately benefit or harm specific subgroups, which is critical in domains such as criminal justice, hiring, healthcare, and education. The dual-objective optimization makes the system capable of balancing raw accuracy with social responsibility, embodying a modern paradigm of ethical machine learning. The architecture of the diagram displayed outlines the end-to-end pipeline used in our study to integrate graph construction, graph-based learning, and fairness-aware optimization within a unified GCN framework.

The use of the *k*-NN procedure involves the integration of node (or vertex) attributes, the computation of feature similarity for neighborhood definition, the addition of self-loops to preserve self-information during aggregation, and the normalization of the adjacency matrix to maintain numerical stability. The combination of all these tasks sets the ground for the graph in its face to learning. The learning process starts with the first GCN layer that gets feature information from each node's direct neighbors. It turns out that the parameter *k* in the *k*-NN algorithm imposes strong effect in the development of an efficient graph structure. To this end, the final step in our inquiry is to study the above effect in terms of statistical inference.

## B. Fairness Analysis Based on Equalized Odds Criterion

Herein, the basic mathematical formulation of the EO fairness metric is described. Let us assume that the available data set consists of $N$ input-output data, where the input data are defined in a $d$-dimensional feature space, and the discrete random variable $S$ that corresponds to the protected attribute,

$$D = \{X, S, Y\} \quad (1)$$

where $X = [x_k]|_{k=1}^{n}$ is the input feature matrix input attributes with $x_k \in \Re^d$, $Y = [y_k]|_{k=1}^{n}$ is the output attribute with $y_k \in \{0,1\}$, and $S = [s_k]|_{k=1}^{n}$ with $s_k$ being equal to 1 if $x_k$ belongs to the privileged group otherwise it is equal to 0. For the random variable $S$ and $Y$ we can also write $S \in \{0,1\}$ and $Y \in \{0,1\}$.

*Definition 1 (Equalized Odds)* [8]. Given the data set $D = \{X, S, Y\}$ and the classifier's predicted output $\hat{Y}$, we say that the classifier satisfies the Equalized Odds (EO) criterion with respect to $S$ and $Y$ if $\hat{Y}$ and $S$ are conditionally independent given $Y$, which implies that $\forall y \in \{0,1\}$ it holds that,

$$P\left(\hat{Y} = 1 \mid Y = y,\ A = 0\right) = P\left(\hat{Y} = 1 \mid Y = y,\ A = 1\right) \quad (2)$$

Note that eq. (2) can be decomposed into the following two equations,

$$P\left(\hat{Y} = 1 \mid Y = 1,\ A = 0\right) = P\left(\hat{Y} = 1 \mid Y = 1,\ A = 1\right) \quad (3)$$

$$P\left(\hat{Y} = 1 \mid Y = 0,\ A = 0\right) = P\left(\hat{Y} = 1 \mid Y = 0,\ A = 1\right) \quad (4)$$

Eq. (3) states that the true positive rates (TPRs) across the two groups defined by the protected attribute should be equal. Relationally, eq. (4) postulates that the FPRs across the above groups should also be equal. Regarding, the condition in eq. (3), whenever the TPRs are equal the same holds for the FNRs, a fact described by the subsequent modification of eq. (3),

$$P\left(\hat{Y} = 0 \mid Y = 1,\ A = 0\right) = P\left(\hat{Y} = 0 \mid Y = 1,\ A = 1\right) \quad (5)$$

We say that the classifier $\hat{Y}$ fulfills the EO criterion if the conditions (4) and (5) are simultaneously satisfied. FPRs and FNRs are also known as type I and II error rates, respectively [9]. Both of them refer to the equality in probability of a person in the unprivileged group being assigned a positive outcome [11]. The importance of eqs (4) and (5) is evident in areas where decisions are affected by the results of a model. Equal FPRs and FNRs provide potential to the model in correctly predicting positive cases. However, when FNRs or FPRs are disproportionate, the model's behavior changes across groups. Especially in the case of FNRs, that situation might obtain increased type II error rates, because high disparity in FNRs implies that instances from different groups have been incorrectly classified. When strong imbalances between the two parts of eqs (4) are present, higher FNR values correspond to the privileged group, whereas lower FNR values are related to stricter predictions of the classifier and correspond to the unprivileged group [12]. Similar results can be identified in the case of FPRs that concerns the type I error rates.

## C. Design of the GCN

This methodology outlines a structured, generalizable approach for applying GCNs to tabular datasets by first transforming them into graph-structured representations using a $k$-NN preprocessing and subsequently integrating fairness-aware learning objectives. The primary goal is to enable accurate and equitable classification while addressing structural and ethical challenges inherent in real-world datasets that often contain sensitive demographic attributes. The process begins with loading a tabular dataset that includes both input features and ground truth outcomes (i.e., target labels) along with at least one protected attribute (e.g., gender, race, or age group). The target label is defined according to the prediction objective, typically as a binary classification task. The protected attribute is mapped to a numeric form and filtered to retain only the groups of interest (e.g., binary groups for comparative fairness analysis).

Uninformative or irrelevant columns (such as IDs, raw labels, or metadata) are dropped to reduce noise and dimensionality. Categorical features are transformed using one-hot encoding to ensure compatibility with numerical learning algorithms. Finally, all features are normalized to a [0, 1] range using MinMax scaling to facilitate more stable and meaningful distance computations during graph construction. Since tabular data does not inherently possess a graph structure, a $k$-NN graph is constructed to define edge connections based on feature similarity. Each data instance is treated as a node, and edges are created between a node and its $k$ most similar neighbors using a chosen distance metric (typically Euclidean or cosine distance). In this study, cosine similarity is adopted as the primary distance metric, as it emphasizes the directional relationship between feature vectors, which is particularly effective in high-dimensional spaces where magnitude may be less informative than orientation. This results to a directed or undirected graph, where the obtained topology reflects local relationships in the original feature space. The resulting edges are encoded into an adjacency format suitable for input into graph-based neural networks. Additionally, node features and labels are converted into tensor representations. This graph-structured data forms the basis for downstream GCN-based learning.

The architecture used is a two-layer GCN, where each layer performs message passing between nodes and aggregates information from local neighborhoods. Fig. 2 provides a conceptual overview of the core architecture of the GCN used in this paper.

The pipeline begins with a graph-structured representation of the input data, where each node corresponds to an individual instance originally derived from a tabular dataset and edges represent pairwise relationships determined via cosine similarity. As mentioned above, the resulting graph structure is based on a $k$-NN preprocessing step that defines the connectivity patterns among nodes. Each node carries an attribute vector in the form of feature representation, where the features are propagated through the graph using successive convolution operations. As a subsequent mode, the first GCN layer performs localized feature aggregation by combining information from each node's neighbors, which assists the model learn not only from the perspective of individual characteristics but also considering contextual relationships.

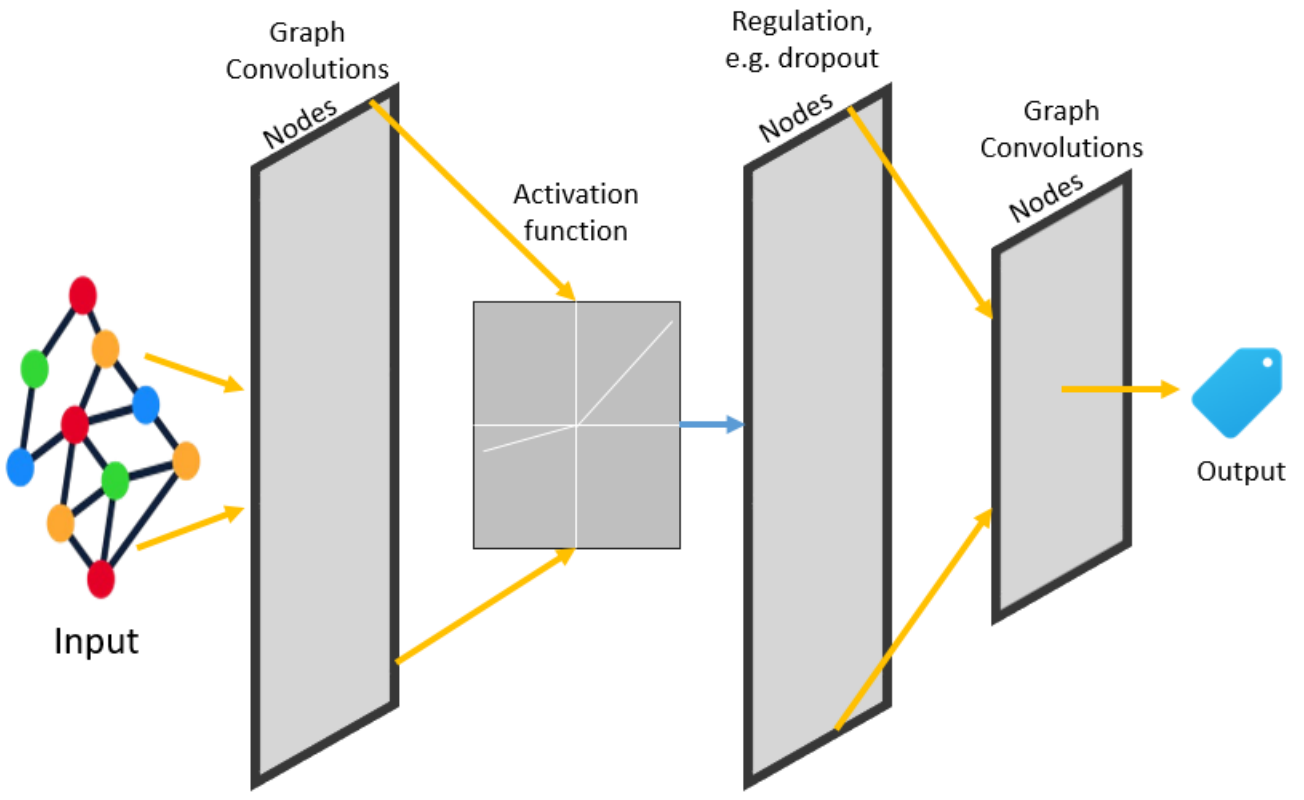

Fig. 2. The basic outline of the used GCN.

After this initial aggregation, the resulting node representations are passed through a non-linear activation function (e.g., ReLU) which allows for learning more complex patterns. A regularization mechanism, such as dropout, is applied in sequence to reduce overfitting by randomly deactivating neurons during training. To this end, a second layer is involved to capture higher-order interactions within the graph. The refined embeddings are then transformed into predictive outputs in terms of a fully connected layer, producing scores or probabilities for each node. Ultimately, the model generates label predictions for the nodes, informed both by involved features and graph-based context. The resulting architecture enables effective learning on structured data, especially in the case when the spatial or relational properties between data instances play a crucial role in model performance.

The GCN operates over graph-structured data, enabling representation learning where both features and relationships (edges) are utilized. The forward pass of a two-layer GCN is mathematically represented as [11]:

$$Z = \text{softmax}\left(\hat{A}\sigma\left(\hat{A}XW^{(0)}\right)W^{(1)}\right) \tag{6}$$

where $X \in R^{n \times d}$ is the input feature matrix for $n$ nodes with $d$ features, $\hat{A} = \tilde{D}^{-1/2}\tilde{A}\tilde{D}^{-1/2}$ is the symmetrically normalized adjacency matrix with self-loops, where $\tilde{D}$ is the degree matrix, a diagonal matrix whose entries $D_{ii}$ represent the sum of edge weights connected to node $i$, including self-loops, and is used to normalize the graph structure, $\tilde{A} = A + I$ is the adjacency matrix with added self-connections, $W^{(0)}W^{(1)}$ are trainable weight matrices, $\sigma(\cdot)$ is the ReLU function, $Z \in R^{n \times c}$ is the output matrix with class scores for each node, where in our case $c = 2$. The above equation define the forward propagation rule of a two-layer GCN [13]. In what follows, more detailed description of the operational status of eq. (6) is provided. The formulation of that equation enables the integration of node attributes and graph topology into the learning process for node-level classification tasks. The input to the model is the node feature matrix $X$. The adjacency matrix $A$ is augmented with self-loops to form $\tilde{A}$ allowing each node to retain its own features during aggregation. This is further normalized to obtain ensuring numerical stability and balanced contribution from neighboring nodes. The model applies a sequence of linear transformations through trainable weight matrices $W^{(0)}$ and $W^{(1)}$ interleaved with the ReLU function. The final output $Z$ represents the class score distribution for each node across the $c$ target classes. This architecture allows GCNs to effectively learn rich, localized node embeddings that capture both structural and attribute-based dependencies within the graph. Since the use of dataset is a lack of an underlying graph structure, we apply $k$-NN to build edges based on feature equality. For each node $i$, the edges are made for its top-to-top neighbors depending on the cosine similarity. Then, the adjacent matrix $A$ is calculated accordingly.

Since datasets lack inherent graph structure (like tabular data), we construct the graph using $k$-NN. In this case, the edge set $E$ is defined as:

$$E = \left\{(i, j) : j \in Top - k\left(sim\left(x_i, x_j\right)\right), i \neq j\right\} \tag{7}$$

where $x_i, x_j \in R^d$ are the feature vectors of the $i$-th and $j$-th nodes, $sim(x_i, x_j)$ is a similarity or distance metric (e.g., cosine similarity), and $Top - k(\cdot)$ selects the $k$ most similar nodes for each node.

The resulting edge list is used to define the adjacency matrix $A$ passed into the GCN. The $k$-NN process defines the information flow between nodes [14]. A small $k$ captures local structure, while a larger $k$ increases graph density, potentially smoothing the resulting graph representations. To this end, the graph is no longer fixed by domain knowledge but constructed from data, meaning graph topology directly affects learning in the GCN.

### D. Optimization Process

A custom loss function is developed to insert fairness constraints during training. Specifically, the loss function used consists of the standard binary cross-entropy that focuses on optimizing the network regarding the classification accuracy, and a fairness-specific part that constrains the above optimization procedure by quantifying the EO criterion. The EO loss penalizes the model for unequal false positive and false negative rates across groups defined by the protected attribute, promoting balanced treatment. The resulting constraints are considered in the optimization scheme in terms of an L1-regularization (i.e., LASSO) regularization approach, which gradually increases the contribution of the fairness term over training epochs to avoid early destabilization of the minimization procedure. Given that the probabilities as calculated by the softmax function are denoted as $p_i = p_i(W)$ $(i = 1, 2, ..., n)$, the binary cross-entropy is,

$$L_{CE}(W) = -\frac{1}{n}\sum_{i=1}^{n}\left(y_i \ln\left(p_i\right) + \left(1 - y_i\right)\ln\left(1 - p_i\right)\right) \tag{8}$$

The FPR condition in eq. (4) is related to the following function [15],

$$h_{FPR}(W)=\left|\frac{\sum_i p_i(1-y_i)s_i}{\sum_i s_i}-\frac{\sum_i p_i(1-y_i)(1-s_i)}{\sum_i(1-s_i)}\right| \quad (9)$$

Relationally, the FNR condition related to eq. (5) can be written as,

$$h_{FNR}(W)=\left|\frac{\sum_i (1-p_i)y_i s_i}{\sum_i s_i}-\frac{\sum_i (1-p_i)y_i(1-s_i)}{\sum_i(1-s_i)}\right| \quad (10)$$

Thus, the constrained optimization problem imposed by the EO criterion as described in eqs (4) and (5) reads as,

$$\begin{aligned} &\text{minimize } L_{CE}(W) \\ &\text{subject to } H_{EO1}(W)=\left(h_{FPR}(W)\right)^2-\delta\leq 0 \\ &\qquad\qquad H_{EO2}(W)=\left(h_{FNR}(W)\right)^2-\delta\leq 0 \end{aligned} \quad (11)$$

Herein, we resolve the above problem in terms of the subsequent regularization approach,

$$L(W)=L_{CE}(W)+\alpha\left(H_{EO1}(W)+H_{EO2}(W)\right) \quad (12)$$

where $\alpha$ is the regularization parameter that controls the counterbalance between the two parts in eq. (12). The total loss function $L(W)$ is minimized in terms of the Adam optimizer. As shown in eq. (12) it consists of two parts namely the task loss function $L_{CE}(W)$ and the fairness loss function $H_{EO}(W)=H_{EO1}(W)+H_{EO2}(W)$ . That customizes the prediction accuracy and fairness where inequalities in mistake rates among sensitive demographic groups are obviously penalized by the model's inclusion of fairness loss in its aim. This method guarantees that the model will continue to perform similarly across a range of those groups once it has learned to accurately identify instances.

## III. Experimental Study

In this section, we provide a systematic experimental evaluation of the methodology regarding the effects of parameter *k*, in the *k*-NN algorithm, considering certain performance metrics. Given that the problem belongs to the realm of binary classification, the metrics used are extracted from the corresponding confusion matrices and include the accuracy (*A*) and the recall (*R*), which are respectively written in terms of true positive (TP), false positive (FP), false negative (FN), and true negative (TN) as follows,

$$A=\frac{TP+TN}{TP+FP+FN+TN} \quad (13)$$

$$R=\frac{TP}{TP+FN} \quad (14)$$

The quantity *R* was selected because it is identical to the TPR and therefore, as mentioned previously, it straightforwardly quantifies the behavior of type II errors (i.e., FNRs).

The study employs three well-known datasets to conduct related experiments. The first is the COMPAS Recidivism dataset [16], where the method attempts to predict the likelihood of criminal re-offense within two years. The protected attribute is the “race”. The second is the Adult Income dataset [17], where the designed network predicts whether an individual earns more than $50K annually. The protected attribute is the “gender”. Finally, the third one is the Law School Admissions dataset reported in [18], where the task is to predict whether a law student will pass the bar exam, based on LSAT, GPA, gender, and race. The protected attribute used is the “gender”. To design the GCN model, each record is transformed into a graph node, where edges represent local similarities in the feature space. That transformation enables the use of GCNs to exploit not only individual characteristics but also relational context among samples [10]. Furthermore, since all three datasets contain well-defined protected attributes and known fairness concerns, they provide an ideal testbed for applying EO regularization during training. In addition, each dataset is prepared to remove irrelevant or highly unbalanced classes and passes through a scaling facility using minmax generalization. Sensitive characteristics and target labels are clearly extracted for later fair assessment.

To evaluate model’s performance and fairness behavior, a systematic experimental protocol was adopted. Specifically, the parameter *k* that were explored were equal to *k* = 3, 5, and 7, where the effect of the resulting graph densities on the learning outcomes were also evaluated. For each combination of dataset and *k*-value, 50 independent runs were conducted to ensure statistical robustness. In every run, 100 epochs were executed allowing for stable convergence of both task-specific and fairness objectives. Prior to training, each dataset was consistently partitioned into a training subset comprising 80% of the instances and a testing subset comprising the remaining 20%, maintaining stratification with respect to the target label to preserve class balance across splits.

### A. *Hypothesis Formulation and Statistical Testing Framework*

This study investigates the impact of the *k*-NN hyperparameter *k* on classification accuracy across the distinct datasets used. While *k*-NN is widely used for its simplicity, the optimal choice of *k* remains dataset-dependent and critically affects performance. The *k*-NN algorithm was implemented with cosine similarity and uniform weighting, controlling for all parameters except *k*.

Table I and Fig. 3 provide the descriptive statistics of the experimental study. In particular, Fig. 3 illustrates the boxplots showing the distribution of classification accuracy across the three different *k*-values for three datasets. In the top panel, corresponding to COMPAS dataset, there is a clear monotonic increase in accuracy as *k* increases, with $k=7$ showing the highest mean accuracy and the narrowest confidence interval, indicating consistent and improved performance with a larger neighborhood size. The middle panel (Adult Income dataset) shows minimal variation in accuracy across all three *k*-values, with overlapping confidence intervals and similar medians, suggesting that model performance is largely independent to the choice of *k*.

The bottom panel (Law School Admissions dataset) reveals a unimodal pattern where $k = 5$ achieves the highest accuracy, while both $k = 3$ and $k = 7$ result in lower and statistically similar performance.

TABLE I. DECSRIPTIVE STATISTICS FOR ALL SIMULATION CASES

| **Dataset** | ***k*-value** | **Mean** | **Median** | **Std** | **Min** | **Max** |
|---|---|---|---|---|---|---|
| COMPAS | *k*=3 | 0.777 | 0.778 | 0.012 | 0.755 | 0.8 |
| | *k*=5 | 0.887 | 0.887 | 0.016 | 0.864 | 0.915 |
| | *k*=7 | 0.920 | 0.919 | 0.008 | 0.909 | 0.936 |
| Adult | *k*=3 | 0.834 | 0.835 | 0.023 | 0.798 | 0.878 |
| | *k*=5 | 0.829 | 0.827 | 0.025 | 0.795 | 0.879 |
| | *k*=7 | 0.840 | 0.84 | 0.009 | 0.822 | 0.854 |
| Law School | *k*=3 | 0.868 | 0.868 | 0.038 | 0.811 | 0.949 |
| | *k*=5 | 0.934 | 0.939 | 0.012 | 0.896 | 0.949 |
| | *k*=7 | 0.867 | 0.868 | 0.038 | 0.811 | 0.947 |

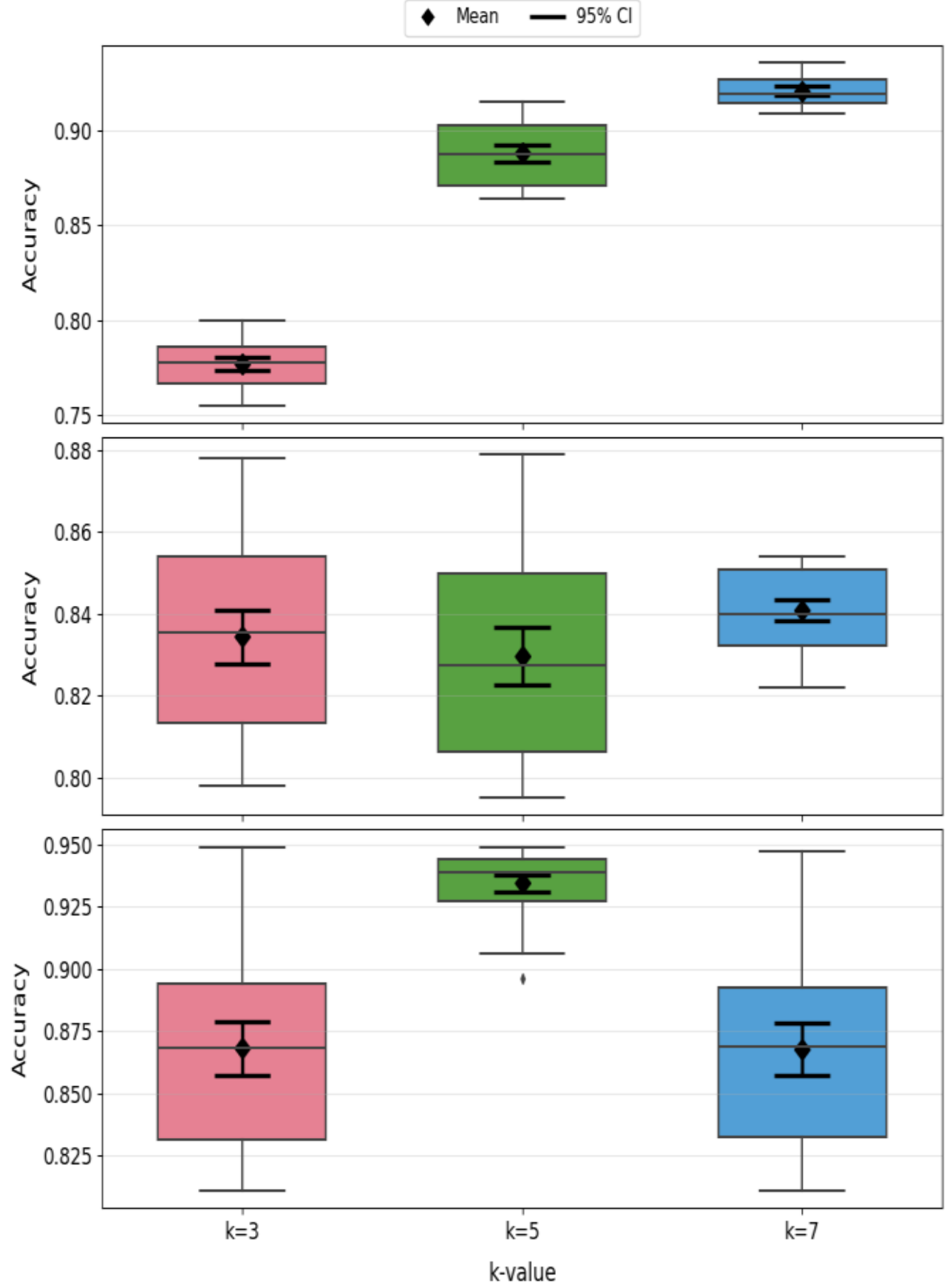

Fig. 3. Box-plots for the comparison of classification accuracy across different *k*-values for the evaluated datasets, where top corresponds to COMPAS, middle to Adult, and bottom to Law School.

The above patterns visually support by the results produced by the statistical inference that follows. For the moment, additional remarks extracted by Table I and Fig. 3 indicate that increasing values of *k* highlight the fact that broader neighborhoods might enrich the graph structure in a way that enhances the GCN's learning capability (e.g., as in COMPAS case). Conversely, the flat trend in Adult Income dataset indicates that the graph constructed from its feature space is robust to variations in neighborhood size, potentially due to more homogeneous data distribution. In Law School Admissions dataset, the peak at $k = 5$ and the wider spread at $k = 7$ suggest that excessive connectivity may introduce noise or reduce discriminative power, while too few connections ($k = 3$) fail to fully capture the underlying structure. Overall, the descriptive statistics emphasize the importance of dataset-specific tuning of the *k* parameter in *k*-NN based graph construction to optimize GCN performance.

TABLE II. STATISTICAL COMPARISON BETWEEN THE RESULTS OBTAINED BY USING THE THREE K-VALUES REGARDING THE THREE DATA SETS

| **Dataset** | **Comparison** | **Test** | **p-value** |
|---|---|---|---|
| COMPAS | $k = 3$ vs $k = 5$ | Wilcoxon-Holm | <0.001 |
| | $k = 3$ vs $k = 7$ | Wilcoxon-Holm | <0.001 |
| | $k = 5$ vs $k = 7$ | Wilcoxon-Holm | <0.001 |
| | — | Friedman | <0.001 |
| Adult | — | Friedman | 0.123011506 |
| Law School | $k = 3$ vs $k = 5$ (corrected) | Wilcoxon-Holm | <0.001 |
| | $k = 3$ vs $k = 7$ (corrected) | Wilcoxon-Holm | 0.369 |
| | $k = 5$ vs $k = 7$ (corrected) | Wilcoxon-Holm | <0.001 |
| | — | Friedman | <0.001 |

In Table II, to further analyze the results, statistical analysis is conducted. We used Friedman tests with Wilcoxon signed-rank post-hoc comparisons and Holm-Bonferroni correction, because there were observed paired observations (same folds across *k*-values) and non-normality in residuals (Shapiro-Wilk $p<0.05$ for 5 out of 9 group-dataset combinations). Effect sizes were calculated using matched-pairs rank-biserial correlation (r) to quantify practical significance beyond p-values.

As in the results of Table I and Fig. 3, the statistical inference used showed that increased *k* values generally improve accuracy ($p<0.001$), with *k*=7 achieving peak performance (0.922 ± 0.008 in COMPAS data set). However, we observe notable exceptions where intermediate *k*=5 outperforms, highlighting the importance of dataset-specific tuning. To conduct the statistical analysis, we evaluated the following hypotheses: (a) Null Hypothesis ($H_0$): No significant differences exist in classification accuracy across *k*-values; (b) Alternative Hypothesis ($H_1$): At least one *k*-value yields significantly different accuracy.

Table II presents the comprehensive statistical comparison of classification accuracy across the three *k*-values (*k*=3, 5, 7) for all datasets. The Friedman test results reveal fundamentally different behaviors across datasets, with COMPAS showing strong monotonic improvement ($\chi^2$=93.96, $p<0.001$), Adult demonstrating no significant differences ($\chi^2$=4.19, p=0.123), and Law School Admissions exhibiting a unimodal pattern peaking at *k*=5 ($\chi^2$=62.80, $p<0.001$). These distinct patterns are visually apparent in the accompanying boxplots (Fig. 3), which show both the central tendencies and dispersion of accuracy measurements.

Building on these findings, recent studies have highlighted that the optimal value of k in *k*-NN-based graph construction can vary significantly depending on the intrinsic structure and feature reparability of the dataset. For instance, in [19] it was shown that in datasets with strong intra-group similarity, larger *k* values enhance information flow, matching the trend seen in COMPAS. Conversely, in datasets where class boundaries are less distinct or the data distribution is more uniform (e.g., Adult Income) larger neighborhoods may introduce noise, diluting useful signals and yielding marginal or non-significant improvements. This reinforces the notion that the interplay between graph connectivity and classification efficacy is highly context-dependent, and that a fixed *k*-value may not be universally optimal across domains.

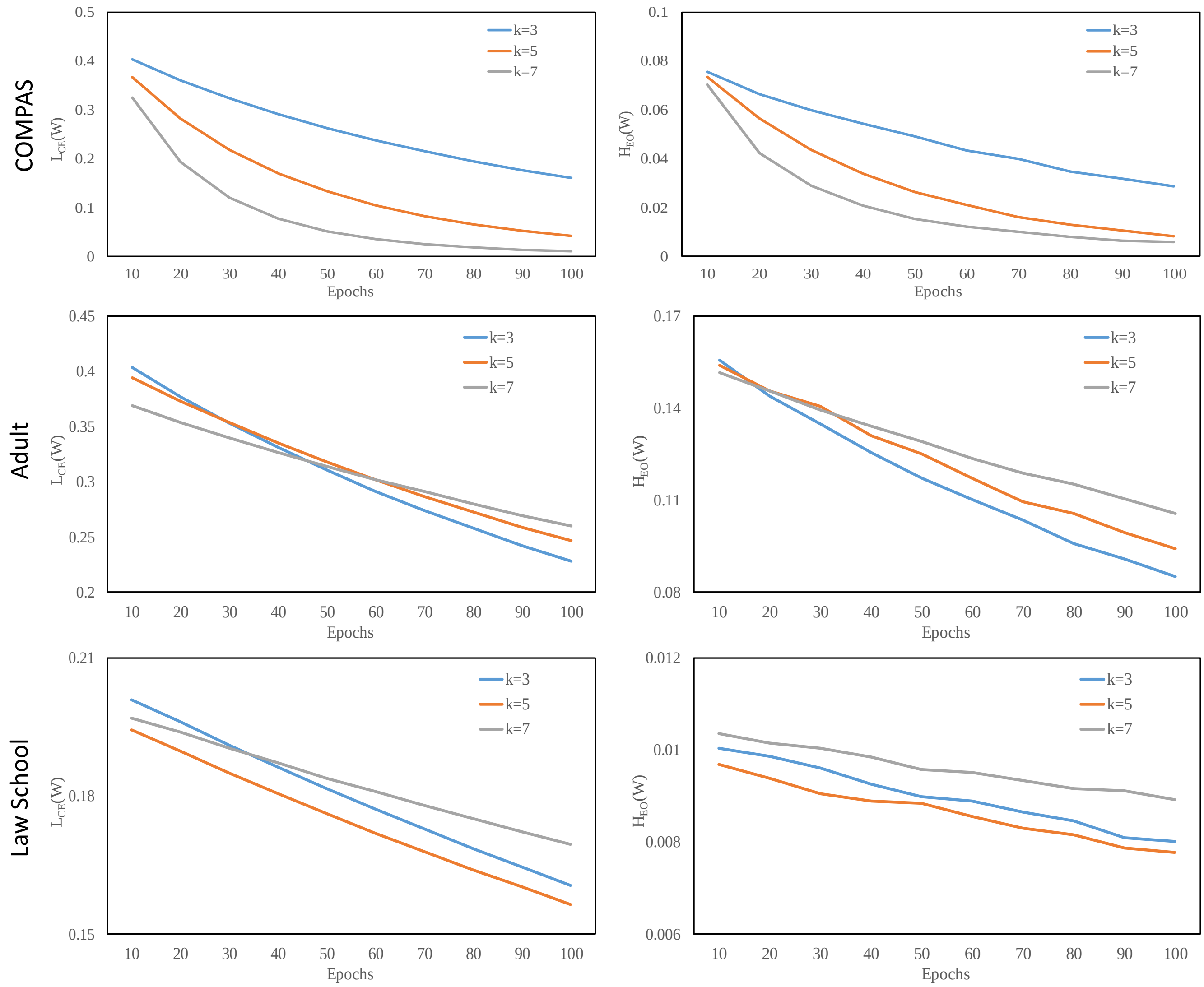

Fig. 4. Evolution of Task Loss and Fairness Loss over training epochs for *k*=3, *k*=5, and *k*=7 across the evaluated datasets.

The hypothesis testing framework yields important insights about parameter sensitivity and model responsiveness to graph structural changes introduced via *k*-NN. For COMPAS and Law School Admissions datasets, we confidently reject the null hypothesis ($H_0$) in favor of the alternative ($H_1$), confirming that *k*-value selection significantly impacts model accuracy. The post-hoc tests reveal this intermediate value significantly outperforms both extremes ($p<0.001$ for both comparisons), while *k*=3 and *k*=7 show no significant difference after multiple test correction. This unimodal pattern suggests Law School dataset exhibits a sweet spot in neighborhood size, where moderate connectivity captures sufficient local structure without introducing excessive noise. The retention of $H_0$ for Adult dataset ($p=0.123$) indicates a fundamentally different behavior where accuracy remains relatively stable across *k*-values.

### B. *Fairness Analysis Using Equalized Odds: Balancing Error Disparities and Model Accuracy*

Utilization of the EO metric during the training process significantly impacts group-specific error rates, particularly in mitigating disparities in the FPR and FNR. Prior to the application of fairness constraints, notable disparities were observed between protected groups for instance, African-American individuals in the COMPAS dataset exhibited higher FPR and FNR compared to Caucasian individuals, revealing inherent systemic biases embedded within the dataset.

By introducing EO regularization during training, these disparities are substantially reduced, leading to more balanced error rates across groups. Such mitigation is crucial not only in high-stakes decision-making systems but also in broader applications to avoid perpetuating biased outcomes. Moreover, while implementing fairness constraints, a trade-off between fairness and model accuracy often emerges. As illustrated in Fig. 4, the statistical results across the three datasets support this observation: in the COMPAS dataset, the task loss function notably decreases from 0.4036 ($k$=3) to 0.1607 ($k$=3) over 100 epochs, while fairness loss function concurrently declines, indicating enhanced fairness. Similar trends are present in the Adult dataset, though the fairness loss function values remain relatively higher, reflecting the complexity of aligning fairness and accuracy in socio-economic data. In the Law School Admissions dataset, both task and fairness loss functions exhibit marginal improvements, suggesting a more stable balance between performance and fairness objectives. Despite minor reductions in classification performance, the gains in fairness, reflected through reduced FPR and FNR gaps, are substantial. This highlights that while fairness interventions may slightly degrade optimal predictive power, they deliver significantly fairer outcomes, promoting ethical and equitable machine learning systems.

While EO was selected as the primary fairness metric due to its balanced consideration of both false positive and false negative rates, other fairness definitions such as DP, or Calibration may provide alternative perspectives, especially when specific fairness aspects are prioritized. Future extensions of this work could explore these metrics to understand their impact and potential trade-offs under the same learning framework. Additionally, the regularization strength of the fairness constraint is governed by a fixed hyperparameter $\alpha$. In our experiments, $\alpha$ was empirically set to ensure a reasonable balance between fairness improvement and classification accuracy. While tuning $\alpha$ is outside the main scope of this work, a sensitivity analysis, e.g., evaluating $\alpha$ across multiple values, could offer deeper insight into the fairness–performance trade-off and remains a valuable direction for further investigation.

## IV. Conclusion and Future Work

This study explored the integration of fairness into GCNs through $k$-NN-based graph construction and constrained optimization. By varying the neighborhood size ($k$ = 3, 5, 7) during preprocessing, we systematically evaluated how graph topology influences classification accuracy and fairness across multiple datasets. Embedding the EO criterion into the optimization objective successfully reduced disparities in false-positive and false-negative rates between sensitive groups, while only marginally affecting overall model accuracy. Scalability remains a concern, but approximate $k$-NN or sparse graphs could mitigate computational costs. These findings emphasize that fairness constraints, when thoughtfully applied, can lead to more equitable outcomes without substantially compromising predictive performance. Though applied to tabular data, the approach may extend to native graphs like social or citation networks. Future work could include MLP baselines with fairness constraints to isolate the graph's impact.

Future work includes adaptive $k$ selection, alternative fairness criteria, and testing on diverse GNNs and large-scale graphs to assess fairness–accuracy trade-offs. This study supports the shift toward responsible and adaptive graph-based learning.

## Acknowledgement

This research was conducted in the context of FAIRPReSONS EU project with no. 101160473, funded within e-JUSTICE program (JUST-2023-JACC-EJUSTICE), (https://fair-presons.aegean.gr/).

## References

[1] Z. Ye, Y. J. Kumar, G. O. Sing, F. Song and J. Wang, "A Comprehensive Survey of Graph Neural Networks for Knowledge Graphs," in *IEEE Access*, vol. 10, pp. 75729-75741, 2022, doi: 10.1109/ACCESS.2022.3191784.

[2] H.-Y. Cai; V. W. Zheng; K. C.-C. Chang. A comprehensive survey of graph embedding: problems, techniques, and applications. IEEE Transactions on Knowledge and Data Engineering, 30(9), 1616-1637, 2018

[3] Munikoti, S., Agarwal, D., Das, L., Halappanavar, M., & Natarajan, B. (2023). Challenges and opportunities in deep reinforcement learning with graph neural networks: A comprehensive review of algorithms and applications. *IEEE transactions on neural networks and learning systems*.

[4] N. Meharbi, F. Morstatter, N. Saxeva, K. Lerman, and A. Galstyan. A survey on bias and fairness in machine learning. arXiv:1908.09635v3 [cs.LG] 25 Jan 2022.

[5] M. Feldman, S. A. Friedler, J. Moeller, and C. Scheidegger, “Certifying and removing disparate impact”, in the Proceedings of the 21st ACM SIGKDD International Conference on Knowledge Discovery and Data Mining (KDD '15), pp. 259–268, 2015.

[6] L. Wu, L. Chen, P. Shao, R. Hong, X. Wang, M. Wang. Learning fair representations for recommendation: A graph-based perspective. arXiv:2102.09140v3 [cs.IR] 23 Apr 2021

[7] J.-Y. Kim, S.-B. Cho. An information theoretic approach to reducing algorithmic bias for machine learning. Neurocomputing 500 (2022) 26–38

[8] M. Hardt, E. Price, and N. Srebro, “Equality of Opportunity in Supervised Learning,” arXiv.org, Oct. 07, 2016. https://arxiv.org/abs/1610.02413v1.

[9] F. Santos, J. Ye, F. Masrour, P. -N. Tan and A. -H. Esfahanian, "FACS-GCN: Fairness-Aware Cost-Sensitive Boosting of Graph Convolutional Networks," *2022 International Joint Conference on Neural Networks (IJCNN)*, Padua, Italy, 2022, pp. 1-8, doi: 10.1109/IJCNN55064.2022.9892919.

[10] Guo, D., Chu, Z., & Li, S. (2023). Fair attribute completion on graph with missing attributes. *arXiv preprint arXiv:2302.12977*.

[11] Ge, H., Dai, Y., Zhu, Z. and Wang, B. (2021) Robust Face Recognition Based on Multi-Task Convolutional Neural Network. Mathematical Biosciences and Engineering, 18, 6638-6651. https://doi.org/10.3934/mbe.2021329.

[12] Yong Shi, Yi Qu, Zhensong Chen, Yunlong Mi, Yunong Wang, Improved credit risk prediction based on an integrated graph representation learning approach with graph transformation, European Journal of Operational Research, Volume 315, Issue 2, 2024, Pages 786-801, ISSN 0377-2217, https://doi.org/10.1016/j.ejor.2023.12.028.

[13] Tamaru, A., Hara, J., Higashi, H., Tanaka, Y., & Ortega, A. (2024, April). Optimizing k in kNN Graphs with Graph Learning Perspective. In *ICASSP 2024-2024 IEEE International Conference on Acoustics, Speech and Signal Processing (ICASSP)* (pp. 9441-9445). IEEE.

[14] Kipf, T. N., & Welling, M. (2016). Semi-supervised classification with graph convolutional networks. *arXiv preprint arXiv:1609.02907*.

[15] P. Manisha, and S. Gujar, “FNNC: Achieving Fairness through Neural Networks”, 2020, arXiv:1811.00247v3. [Online]. Available: https://arxiv.org/abs/1811.00247.

[16] *COMPAS Recidivism Dataset*. Kaggle. Retrieved from https://www.kaggle.com/datasets/danofer/compass.

[17] *Adult Income Dataset*. UCI Machine Learning Repository. Retrieved from https://archive.ics.uci.edu/dataset/2/adult.

[18] Law School Admissions – Bar Passage Dataset. Kaggle. Retrieved from https://www.kaggle.com/datasets/danofer/law-school-admissions-bar-passage/data.

[19] S. Zhou *et al*., "A Method to Automatic Create Dataset for Training Object Detection Neural Networks," in *IEEE Access*, vol. 10, pp. 80505-80517, 2022, doi: 10.1109/ACCESS.2022.3195490.